\documentclass{article}

\usepackage{arxiv}

\usepackage[utf8]{inputenc} 
\usepackage[T1]{fontenc}    
\usepackage{hyperref}       
\usepackage{url}            
\usepackage{booktabs}       
\usepackage{amsfonts}       
\usepackage{nicefrac}       
\usepackage{microtype}      
\usepackage{lipsum}
\usepackage{amsmath}
\usepackage{amssymb}
\usepackage{graphicx}
\usepackage[makeroom]{cancel}
\graphicspath{ {./images/} }

\title{A Brief Review of Domain Adaptation}

 \author{
    Abolfazl Farahani \\
    Department of Computer Science\\
    University of Georgia\\
    Athens, GA, USA \\
    \texttt{a.farahani@uga.edu} \\
    \And
    Sahar Voghoei \\
    Department of Computer Science\\
    University of Georgia\\
    Athens, GA, USA \\
    \texttt{Voghoei@uga.edu} \\
    \And
    Khaled Rasheed \\
    Department of Computer Science\\
    University of Georgia\\
    Athens, GA, USA \\
    \texttt{Khaled@uga.edu} \\
    \And
    Hamid R. Arabnia \\
    Department of Computer Science\\
    University of Georgia\\
    Athens, GA, USA \\
    \texttt{hra@uga.edu} \\
  }

\begin{document}
\maketitle
\begin{abstract}
Classical machine learning assumes that the training and test sets come from the same distributions. Therefore, a model learned from the labeled training data is expected to perform well on the test data. However, This assumption may not always hold in real-world applications where the training and the test data fall from different distributions, due to many factors, e.g., collecting the training and test sets from different sources, or having an out-dated training set due to the change of data over time. In this case, there would be a discrepancy across domain distributions, and naively applying the trained model on the new dataset may cause degradation in the performance. Domain adaptation is a sub-field within machine learning that aims to cope with these types of problems by aligning the disparity between domains such that the trained model can be generalized into the domain of interest. This paper focuses on unsupervised domain adaptation, where the labels are only available in the source domain. It addresses the categorization of domain adaptation from different viewpoints. Besides, It presents some successful shallow and deep domain adaptation approaches that aim to deal with domain adaptation problems.
\end{abstract}


\section{Introduction}
    Traditional Machine learning aims to learn a model on a set of training samples to find an objective function with minimum risk on unseen test data. To train such a generalized model, however, it assumes that both training and test data are drawn from the same distribution and share similar joint probability distributions. This constraint can be easily violated in real-world applications since training, and test sets can originate from different feature spaces or distributions.
    The difficulty of collecting new instances with the same property, dimension, and distribution, as we have in the training data may happen due to various reasons, e.g., the statistical properties of a domain can evolve in time, or new samples can be collected from different sources, causing domain shift. Besides, if possible, it is preferred to utilize a related publicly available annotated data as a training dataset instead of creating a new labeled dataset, which is a laborious and time-consuming task. However, when the training data is not an accurate reflection of test data distribution, the trained model will likely experience degradation in performance when applying on the test data. To tackle the above problem, researchers proposed a new research area in machine learning called domain adaptation. 
    In this setting, training and test sets termed as the source and the target domains, respectively. Domain adaptation generally seeks to learn a model from a source labeled data that can be generalized to a target domain by minimizing the difference between domain distributions.
    
    Domain adaptation is a special case of transfer learning \cite{pan2010survey}. These two closely related problem settings are sub-discipline of machine learning which aim to improve the performance of a target model with insufficient or lack of annotated data by using the knowledge from another related domain with adequate labeled data.
    We first briefly review transfer learning and its categories, and then address some existing related works to transfer learning and domain adaptation.
    Transfer learning refers to a class of machine learning problems where either the tasks and/or domains may change between source and target while in domain adaptations only domains differ and tasks remain unchanged. In transfer learning, a domain consists of feature space and marginal probability distribution, and a task includes a label space and an objective predictive function. Thus, Various possible scenarios in domains and tasks create different transfer learning settings.
    Pan \textit{et al.} \cite{pan2010survey}, categorizes transfer learning into three main categories: inductive, transductive, and unsupervised transfer learning.  
    Inductive transfer learning refers to the situation where the source and target tasks differ, no matter whether or not domains are different. In this setting, the source domain may or may not include annotated data, but a few labeled data in the target domain are required as training data.
    In transductive transfer learning, tasks remain unchanged while domains differ, and labeled data are available only in the source domain. However, part of the unlabeled data in the target domain is required at training time to obtain its marginal probability distribution.
    Finally, unsupervised transfer learning refers to the scenario where the tasks are different similar to inductive transfer learning; however, both source and target domains include unlabeled data.
    
    Similar to transfer learning and domain adaptation, 
    Semi-supervised classification addresses the problem of having insufficient labeled data. This problem setting employs abundant unlabeled samples and a small amount of annotated samples to train a model. In this approach, both labeled and unlabeled data samples are assumed to be drawn from the equivalent distributions. In contrast, transfer learning and domain adaptation relax this assumption to allow domains to be from distinct distributions \cite{zhu2005semi}, \cite{pise2008survey}.
    
    Multi-task learning \cite{caruana1997multitask}, is another related task that aims to improve generalization performance in multiple related tasks by simultaneously training them. Since related tasks are assumed to utilize common information, Multi-task learning tends to learn the underlying structure of data and share the common representations across all tasks. There are some similarities and differences between transfer learning and multi-task learning. Both of these learning techniques use similar procedures, such as parameter sharing and feature transformation, to leverage the transferred knowledge to improve learners' performance. However, Transfer learning aims to boosts the target learner performance by first training a model on a source domain and then transferring the related knowledge to the target learner, while multi-task learning aims to improve the performance of multiple related tasks by jointly training them.
    
    Similarly, multi-view learning aims to learn from multi-view data or multiple sets of distinctive features such as audio+video, image+text, text+text, etc. E.g., a web page can be described by the page and hyperlink contents, which is an example of describing data by two different sources of text. The intuition behind this type of learning is that the multi-view data contains complementary information, and a model can learn more comprehensive and compact representation to improve the generalization performance.
    Some real-world application examples of multi-view learning are video analysis \cite{ramanishka2016multimodal}, \cite{tran2015learning}, speaker recognition \cite{livescu2009multi}, \cite{asali2020deepmsrf}, natural language processing \cite{cho2014learning}, \cite{rush2015neural}, \cite{serban2016building}, and recommender system \cite{wang2011collaborative}, \cite{wang2015collaborative}. 
    Canonical correlation analysis (CCA) \cite{hotelling1992relations}, and co-training \cite{blum1998combining}, are the first representative techniques introduced in the concept of multi-view learning which are also used in transfer learning \cite{bo2015transfer}.
    
    Domain generalization \cite{muandet2013domain}, also tends to train a model on multiple annotated source domains which can be generalized into an unseen target domain. In domain generalization, target samples are not available at the training time. However, domain adaptation requires the target data during training to align the shift across domains.
\section{Notations and definitions}
    In domain adaptation, domains can be considered as three main parts: input or feature space $\mathcal{X}$, output or label space $\mathcal{Y}$, and an associated probability distribution $p(x,y)$, i.e., $\mathcal{D}=\{\mathcal{X},\mathcal{Y}, p(x,y)\}$. Feature space $\mathcal{X}$ is a subset of a d-dimensional space, $\mathcal{X}\subset \mathbb{R}^d$, $\mathcal{Y}$ refers to either a space of binary $\{-1,+1\}$ or multi-class $\{1,...K\}$, where $K$ is the number of classes, and $p(x,y)$ is a joint probability distribution over the feature-label space pair $\mathcal{X} \times \mathcal{Y}$. We can decompose the joint probability distribution as $p(x,y)= p(x)p(x|y)$ or $p(x,y)=p(y)p(y|x)$, where $p(.)$ is a marginal distribution and $p(.|.)$ is a conditional distribution.
    
    Given a source domain $\mathcal{S}$ and a target domain $\mathcal{T}$, the source dataset samples drawn from the source domain consist of feature-label pairs $\{(x_i, y_i)\}_{i=1}^n$, where $n$ is the number of samples in the source data set. Similarly, the target data set can be denoted as $\{z_i, u_i\}_{i=1}^m$, where $(z_i,u_i)$ refers to the target samples and their associated labels. In unsupervised domain adaptation where the labels are not available in the target domain, $u$ is unknown. 
    When the source and target domains are related but from different distributions, naively extending the underlying knowledge contained in one domain into another might negatively affect the learner's performance in the target domain. 
    Therefore, domain adaptation was proposed to tackle this problem by reducing the disparity across domains and further training a model that performs well on the target samples. In other words, domain adaptation aims to learn a generalized classifier in the presence of a shift between source and target domain distributions.
    Classification is a machine learning task that aims to learn a function from labeled training data to map input samples to real numbers $h:\mathcal{X}\xrightarrow{}\mathcal{Y}$, where $h$ is a function or an element of a hypothesis space $\mathcal{H}$, and $\mathcal{H}$ refers to a set of all possible functions. For example, in image classification, a classifier assigns each input image a category such as a dog or a cat.
    Generally, to obtain the best predictive function, we learn a model on a given source dataset by minimizing the expected risk of the source labeled data:
    \begin{equation}
    \begin{array}{rcl}
    R_S(h)&=\mathbb{E}_{(x,y)\sim P_S(x,y)}\left[\ell(h(x),y)\right]\\
        &=\sum\limits_{y\in \mathcal{Y}}\int_\mathcal{X}\ell(h(x),y)p_S(x,y)dx,
    \end{array}
    \end{equation}
    where the expectation is taken with respect to the source distribution $P_S$, $\ell(h(x),y)$ is a loss function that denotes the disagreement between the label predicted by the classifier $h$ and the true label, and $R_S(h)$ is the sum of all the disagreements or misclassified samples in the source domain. However, in the supervised learning, the goal is to learn a model with the lowest risk when applying on the target domain. Thus, we can rewrite the above equation as follows:
    \begin{equation}
    \begin{array}{rcl}
    R_\mathcal{T}(h)&=& \mathbb{E}_{(x,y)\sim P_T}\left[\ell(h(x),y)\right]\\
                    &=& \sum\limits_{y\in \mathcal{Y}}\int_\mathcal{X}\ell(h(x),y)p_\mathcal{T}(x,y)dx\\
                    &=& \sum\limits_{y\in \mathcal{Y}}\int_\mathcal{X}\ell(h(x),y)p_\mathcal{T}(x,y)\frac{p_\mathcal{S}(x,y)}{p_\mathcal{S}(x,y)}dx\\
                    &=& \sum\limits_{y\in \mathcal{Y}}\int_\mathcal{X}\ell(h(x),y)\frac{p_\mathcal{T}(x,y)}{p_\mathcal{S}(x,y)}p_\mathcal{S}(x,y)dx\\
                    &=& \mathbb{E}_{(x,y)\sim P_S}\left[\frac{p_\mathcal{T}(x,y)}{p_\mathcal{S}(x,y)}\ell(h(x),y)\right],
    \end{array}
    \end{equation}
    where $P_S(x,y)$ and $P_T(x,y)$ are joint probability distribution of source and target domains. For more information about the risk minimization and supervised learning, see \cite{mohri2018foundations}\cite{vapnik1992principles}.
    In classical machine learning, $\frac{P_T(x,y)}{P_S(x,y)}=1$ since the assumption is that both training and test data are drawn from the same distribution. However, domain adaptation relaxes this assumption, which will be discussed in the following sections.
\section{Categorization of Domain Adaptation}
    Conventional domain adaptation assumes that feature and label spaces remain unchanged while their probability distributions may vary between domains. However, finding a source and target domains with equivalent label space is usually arduous or even impossible. When the source and target label spaces are not identical, matching the whole source and target distributions will create a representation space containing the features of the data belonging to the source classes which do not exist in the target classes. The target domain sees these classes as outlier classes, and extending their knowledge into the target domain will cause negative transfer, which significantly harms the model performance.
    Thus, in addition to marginal distribution disparity, we need to consider different scenarios where label spaces differ across domains. Different marginal distributions and different label spaces across domains are termed as domain gap and category gap, respectively.
    Based on the category gap, domain adaptation can be divided into four main categories; closed set, open set, partial, and universal domain adaptation.
    \begin{itemize}
        \item Closed set domain adaptation refers to the situation where both source and target domains share the same classes while there still exists a domain gap between domains. Traditional domain adaptation falls into this category.
        \item In open set domain adaptation, related domains share some labels in the common label set and also they may have private labels \cite{panareda2017open}. Saito \textit{et al.} \cite{saito2018open}, introduced new open set domain adaptation in which the data in the source private classes is removed. In the modified open set, the source label set is considered to be a subset of the target label set. 
        Open set domain adaptation is suitable when there are multiple source domains where each includes a subset of target classes. Domain adaptation techniques aim to utilize all the source domain information contained in the shared classes to boost the model's performance in the target domain.
        \item In contrast to open set, partial domain adaptation refers to the situation where the target label set is a subset of the source label set \cite{cao2018partial}, \cite{zhang2018importance}. In this setting, the available source domain can be considered as a generic domain that consists of an abundant number of classes, and the target is only a subset of the source label set with fewer classes.
        \item Universal domain adaptation (UDA) \cite{you2019universal}, generalizes the above scenarios. In contrast to the above settings, which require prior knowledge about the source and target label sets, universal domain adaptation is not restricted to any prior knowledge. In this setting, source and target domains may share common label sets, and also each domain may have a private label set or outlier classes. Universal domain adaptation first tends to find the shared label space across domains and then similar to open set, and partial domain adaptation aligns the data distributions in the common label set. Ultimately, a classifier will be trained on the matched source labeled data to be applied safely to the unlabeled target data. In the testing phase, the trained classifier in both open set and universal domain adaptation is expected to assign accurate labels to the target samples belonging to the shared label space, and mark the samples in the outlier classes as unknown.
    \end{itemize}
    This overview focuses on closed set unsupervised domain adaptation. This type of domain adaptation aims to utilize the labeled source data and the unlabeled target data to learn an objective predictive function that can perform well on the target domain where there is a shift between domains.
    Based on the definition of a domain (see section 2), two domains can be different if at least one of input space, output space, or the probability density function changes between domains. Closed set domain adaptation considers the situation where the feature and label spaces are identical between domains, while the joint probability distributions may differ. Domain shift mainly can be categorized into three classes: prior shift, covariate shift, and concept shift.
    \begin{itemize}
        \item Prior shift or class imbalance considers the situation where posterior distributions are equivalent, $p_s(y|x)=p_t(y|x)$, and prior distributions of classes are different between domains, $p_s(y)\neq p_t(y)$. To solve a domain adaptation problem with a prior shift, we need labeled data in both source and target domains.
        \item Covariate shift refers to a situation where marginal probability distributions differ, $p_s(x)\neq p_t(x)$, while conditional probability distributions remain constant across domains, $p_s(y|x)=p_t(y|x)$. Sample selection bias and missing data are two causes for the covariate shift. Most of the proposed domain adaptation techniques aim to solve this class of domain gap.
        \item Concept shift, also known as data drift, is a scenario where data distributions remain unchanged, $p_s(x)=p_t(x)$, while conditional distributions differ between domains, $p_x(y|x)\neq p_t(y|x)$. Concept shift also requires labeled data in both domains to estimate the ratio of conditional distributions.
        \end{itemize}
\section{Approaches}
    Existing domain adaptation approaches can be broadly categorized into methods with shallow and deep architectures. Shallow domain adaptation approaches \cite{gretton2009covariate}, \cite{gopalan2011domain}, \cite{pan2010domain}, \cite{jhuo2012robust}, mainly utilize instance-based and feature-based techniques to align the domain distributions. One way of aligning the distributions is by minimizing the distance between domains. The mostly used distance measures in domain adaptation are maximum mean discrepancy (MMD) \cite{gretton2007kernel}, Wasserstein metric, correlation alignment (CORAL)\cite{sun2017correlation}, Kullback-Leibler (KL) divergence \cite{kullback1951information}, and contrastive domain discrepancy (CDD) \cite{kang2019contrastive}.
    Deep domain adaptation approaches \cite{long2015learning}, \cite{glorot2011domain}, \cite{ganin2015unsupervised}, \cite{ajakan2014domain}, on the other hand, utilize neural networks. This class of approaches usually use convolutional, autoencoder, or adversarial based networks to diminish the domain gap. Some of the approaches in this category may also utilize a distance metric in one or multiple layers of two networks, one for source data and one for target data, to measure the discrepancy between the feature representations on the corresponding layers. 
    \subsection{Instance-Based Adaptation}
        Instance-based domain adaptation approaches aim to deal with the shift between data distributions by minimizing the target risk based on the source labeled data.   
        As mentioned in section 3, domain shift is mainly categorized in to covariate shift, prior shift and concept shift. 
        In unsupervised domain adaptation where the labeled data is available only in the source domain, If the source and target distributions are different, $p_{\mathcal{T}}(x, y)\neq p_\mathcal{S}(x, y)$, and the source and the target posteriors are arbitrary, $p_\mathcal{T}(y|x)\neq p_\mathcal{S}(y|x)$, the problem in Eq.2 becomes intractable without labeled data in the target domain. Hence, we need to assume that both distributions are different only in marginal distribution while the posteriors remain unchanged. This setting is called covariate shift in which the target risk can be simplified as follows:
        \begin{equation} \label{eq3}
                R_\mathcal{T}(h) = \mathbb{E}_{(x,y)\sim P_S}\left[\frac{p_\mathcal{T}(x)\cancel{p_\mathcal{T}(y|x)}}{p_\mathcal{S}(x)\cancel{p_\mathcal{S}(y|x)} }\ell(h(x),y)\right],
        \end{equation}
        where the ratio of two density functions is considered as importance weight, i.e., $w(x)=\frac{p_\mathcal{T}(x)}{p_\mathcal{S}(x)}$.
        When domains consist of prior shift, the assumption is that the conditional distributions remain equal while prior distributions of classes differ across domains. Therefore, we can simplify the target risk as follows:
        \begin{equation} \label{eq4}
                R_\mathcal{T}(h) = \mathbb{E}_{(x,y)\sim P_S}\left[\frac{p_\mathcal{T}(y)\cancel{p_\mathcal{T}(x|y)}}{p_\mathcal{S}(y)\cancel{p_\mathcal{S}(x|y)} }\ell(h(x),y)\right],
        \end{equation}
        where $w(y)=\frac{p_\mathcal{T}(y)}{p_\mathcal{S}(y)}$ is known as class weights. To solve the prior shift, we need labeled data in both source and the target domains which is out of our scope.
        
        A typical solution to the covariate shift problem is to use importance weighting approaches to compensate for the bias by re-weighting the samples in the source domain based on the ratio of target and source domain densities, $w(x)= p_\mathcal{T}(x)/p_\mathcal{S}(x)$, where $w(x)$ is re-weighting factor for the samples in the source domain.
        \cite{shimodaira2000improving}, \cite{zadrozny2004learning}, proved that using the density ratio could remove sample selection bias by re-weighting the source instances. Thus, this problem can be viewed as a density ratio estimation (DRE). In this approach, the key idea is to find appropriate weights for the source samples through an optimization problem, such that the discrepancy between the re-weighted source data distribution and the actual target data distribution can be minimized.
        
        To estimate the importance, one can indirectly estimate the marginal data distributions of each domain separately and then estimate the ratio. However, indirect density estimation is usually ineffective and a very challenging task, as the importance is usually unknown in reality,  especially when having high dimensional features \cite{hardle2012nonparametric}.
        A solution to this problem is to estimate the weights directly in an optimization procedure in which the model minimizes the discrepancy between the weighted source and target distributions.
        The disparity between domains can be reduced using Kernel Mean Matching (KMM)\cite{huang2007correcting}, \cite{gretton2009covariate}. KMM is a non-parametric method that directly estimates the weights by minimizing the discrepancy between domains using Maximum Mean Discrepancy (MMD) in the Reproducing Kernel Hilbert Space (RKHS). KMM first maps the samples in both domains into RKHS using a nonlinear kernel function such as Gaussian, and then obtains the sample weights by minimizing the means of target data and the weighted source data.
        \begin{equation}
            \begin{array}{c c}
                D_{MMD}[w, p_\mathcal{S}(x), p_\mathcal{T}(x)] = \min\limits_{w}\Vert\mspace{3mu}\mathbb{E}_\mathcal{S}[w(x)\phi(x)]-\mathbb{E}_\mathcal{T}[\phi(x)]\mspace{3mu}\Vert_\mathcal{H}\\
                s.t.\quad w(x)\in[0,W] ,\quad \mathbb{E}_\mathcal{S}[w(x)]=1,
            \end{array}
        \end{equation}
        where $\phi(x)$ is a non-linear kernel function that maps the data samples into RKHS, and both constraints on $w(x)$ ensure that the variance of sample weights are bounded to be low and the weighted source data distribution is close to the probability distribution. Similar to many kernel based approaches, KMM uses quadratic program in the optimization process which restrict the model to work well only on small datasets.
        Kullback-Leibler Importance Estimation Procedure (KLIEP) \cite{sugiyama2005model}, \cite{sugiyama2007covariate}, \cite{sugiyama2008direct}, directly estimates the density ratio using Kl-divergence between target distribution and weighted source distribution. In this setting, KL-divergence can be simplified as follows:
        \begin{equation}
            \begin{split}
                D_{KL}[w(x), p_\mathcal{S}(x), p_\mathcal{T}(x)] & =     \int_\mathcal{X}p_\mathcal{T}(x)log\frac{p_\mathcal{T}}{p_\mathcal{S}w(x)}dx\\
                & = \int_\mathcal{X}p_\mathcal{T}(x)log\frac{p_\mathcal{T}}{p_\mathcal{S}}dx\mspace{5mu} - \int_\mathcal{X}p_\mathcal{T}(x)\log w(x)dx\\
                & \varpropto\quad m^{-1} \sum\limits_{j}^{m}\log w(z_j).
            \end{split}
        \end{equation}
        In the second line of above equation, the first term does not depend on $w(x)$ which means that it is constant and can be removed from the objective function since we are optimizing it w.r.t $w(x)$.
        The above objective function is based on the weighted samples in the target domain that makes it computationally expensive for large scale problems since we need to compute new weights for each new target data. To solve this problem, \cite{sugiyama2005input}, proposed a new model that uses a linear model $w(x)= \phi(x)\alpha $, where $\phi(x)$ is a basis function and $\alpha$ is a set of parameters to be learned. Therefore, the objective function can be written as:
        \begin{equation} \label{eq1}
                D_{KL}[\phi(x), p_\mathcal{S}(x), p_\mathcal{T}(x)] =  m^{-1} \sum\limits_{j}^{m}\log \phi(z_j)\alpha.
        \end{equation}
        KLIEP minimizes the KL-divergence between the actual target distribution and the importance-weighted source distribution in a non-parametric manner to find the instance weights.
    \subsection{Feature-Based Adaptation}
        Feature-based adaptation approaches aim to map the source data into the target data by learning a transformation that extracts invariant feature representation across domains. They usually create a new feature representation by transforming the original features into a new feature space and then minimize the gap between domains in the new representation space in an optimization procedure, while preserving the underlying structure of the original data. 
        Subspace-based, transformation-based, construction-based are some of the main feature-based adaptation methods. Below, we elaborate on each category and address some related approaches.
        \subsubsection{Subscpace-based.}
            Subspace-based adaptation aims to discover a common intermediate representation that is shared between domains. Many techniques have been proposed to construct this representation from the low-dimensional representation of the source and target data.  
            Most of the adaptation approaches in this category first create a low-dimensional representation of original data in the form of a linear subspace for each domain, and then reduce the discrepancy between the subspaces to construct the intermediate representation. A dimensionality reduction technique such as principal component analysis (PCA) can be used to construct the subspaces as two points, one for each domain, in a low-dimensional Grassmann manifold. The distance between the points in Grassmann manifold indicates the domain shift which can be reduced by applying different methods.
            Goplan \textit{\textit{et al.}} \cite{gopalan2011domain}, proposed Sampling Geodesic Flow (SGF) that first finds a geodesic path between the source and target points On a Grassmann manifold and then samples a set of points, subspaces, including the source and the target points along this path. In the next step, the data from both domains are projected onto all sampled subspaces along the geodesic path and will be concatenated to create a high-dimensional vector. Finally, A discriminative classifier can learn from the source projected data to classify the unlabeled samples. Sampling more points from the geodesic path would help to map the source subspace into the target subspace more precisely. However, sampling more subspaces extends the dimensionality of the feature vector, which makes this technique computationally expensive.
            Geodesic Flow Kernel \cite{gong2012geodesic}, was proposed to extend and improve SGF. GFK is a kernel-based domain adaptation method that deals with shift across domains. It aims to represent the smoothness of transition from a source to a target domain by integrating an infinite number of subspaces to find a geodesic line between domains in a low-dimensional manifold space.
            Fernando \textit{et al.} \cite{fernando2013unsupervised}, proposed a subspace alignment technique (SA) to directly reduce the discrepancy between domains by learning a linear mapping function that projects the source point directly into the target point in the grassmann manifold. The projection matrix $M$ can be learned by minimizing the divergence in the Bergman matrix:
            \begin{equation}
                M=\underset{M}{\mathrm{argmin}}\quad||X_\mathcal{S}M-X_\mathcal{T}||_F^2=X_\mathcal{S}^TX_\mathcal{T},
            \end{equation}
            where $X_S, X_T$ are the low-dimensional representation, basis vectors, of The source and the target data in the Grassmann manifold respectively, and $||.||_F^2$ is the Frobenius norm. 
            SA only aligns the subspace bases and ignores the difference between subspace distributions.
            Subspace distribution alignment (SDA) \cite{sun2015subspace}, extends the work in SA by aligning both subspace distributions and the bases at the same time. In SDA the projection matrix $M$ can be formulated as \break $M=X_\mathcal{S}^T X_\mathcal{T}Q$, where $Q$ is a matrix to align the discrepancy between distributions. 
        \subsubsection{Transformation-based.}
            Feature transformation transforms the original features into a new feature representation to minimize the discrepancy between the marginal and the conditional distributions while preserving the original data's underlying structure and characteristics. To reduce the domain discrepancy, we need some metrics such as Maximum Mean discrepancy (MMD), Kullback-Leibler Divergence (KL-divergence), or Bregman Divergence to measure the dissimilarity across domains.
            Transfer component adaptation (TCA) \cite{pan2010domain}, was proposed to learn a domain-invariant feature transformation in which the marginal distributions between the source and target domains are minimized in RKHS using maximum mean discrepancy (MMD) criterion.  After finding the domain-invariant features, we can utilize any classical machine learning technique to train the final target classifier.
            Joint domain adaptation (JDA) \cite{long2013transfer}, extends TCA by simultaneously matching both marginal and conditional distributions between domains. JDA utilizes PCA as a dimensionality reduction technique to extract more robust features. Low-dimensional features can then be embedded into a high-dimensional feature space where the difference between marginal distributions are minimized using MMD. To align conditional distributions, we need labeled data in both domains. When labels are unavailable in the target domain, we can use pseudo labels, which can be estimated by the classifier trained on the labeled source data. After obtaining the pseudo labels, the model minimizes the distance between conditional distributions by modifying MMD. All the above steps are jointly and iteratively performed to find the best mapping function that aligns both marginal and conditional distributions across domains. The final target classifier can be trained on domain-invariant features discovered by the algorithm.
        \subsubsection{Reconstruction-based.}
            The feature reconstruction-based methods aim to reduce the disparity between domain distributions by sample reconstruction in an intermediate feature representation.
            Jhuo \textit{et al.} \cite{jhuo2012robust}, proposed a Robust visual Domain Adaptation with Low-rank Reconstruction (RDALR) method to reduce the domain discrepancy. RDALR learns a linear projection matrix $W$ that transforms the source samples into an intermediate representation where they can be linearly represented by the samples in the target domain to align the domain shift. The domain adaptation problem can be addressed by minimizing the following objective function:
            \begin{equation}
                \begin{array}{ccl}
                    \underset{W,Z,E}{\text{min}}rank(Z)+\alpha ||E||_{2,1},\\
                    s.t. \quad WX_S=X_TZ+E,\quad
                    WW^T=I,
              \end{array}
            \end{equation}
            where $X_S$ and $X_T$ are sets of samples in the source and target domains respectively, $WX_\mathcal{S}$ denotes the transformed source samples, Z is the reconstruction coefficient matrix including a set of coefficient vectors corresponding to the projected source samples, $E$ is a matrix of noise and outlier information of source domain, and $\alpha$ is the regularization parameter. By minimizing the rank of the coefficient matrix $Z$, the method tends to reconstruct the different source samples together, and find the underlying structure of source samples. $WW^T=I$ ensures to have a basis transformation matrix by enforcing it to be orthogonal. Besides, the noises and outliers in the source domain are decomposed into $E$. By Minimizing $E$, the noises and outliers will be removed from the projected source data. However, RDALR is restricted to the rotation only, and the discriminative source domain information may not be transferred to the target domain, causing an unreliable alignment.
            Shao \textit{et al.} \cite{shao2014generalized}, proposed Low-Rank Transfer Subspace Learning (LTSL) to resolve the RDALR problem. LTSL intends to discover a common low-rank subspace between domains where the source samples can reconstruct the target samples. Learning the common subspace makes LTSL more flexible on data representation. LTSL performs the adaptation by minimizing the following objective function:
            \begin{equation}
                \begin{array}{ccl}
                    \underset{W,Z,E}{\text{min}} F(W,X_S)+\lambda_1rank(Z)+\lambda_2 ||E||_{2,1},\\
                    s.t. \quad W^TX_T=W^TX_SZ+E,\quad 
                    W^TU_2W=I,
               \end{array}
            \end{equation}
            where F(.) is a generalized subspace learning function. In contrast to RDALR that considers all target samples to represent the projected source samples, LTSL assumes that each sample in the target domain is more related to its neighborhood. Thus, each datum in the target domain can be reconstructed by only a set of source samples. In this way, the method can transfer both locality and discriminative properties of the source domain into the target domain.
    \subsection{Deep Domain Adaptation}
        In recent years, deep neural networks have widely employed for representation learning and achieved remarkable results in many machine learning tasks such as image classification \cite{hu2018squeeze}, sentiment analysis \cite{zhang2018deep}, speech recognition \cite{dahl2011context}, \cite{deng2010binary}, object detection \cite{ouyang2015deepid}, \cite{zhao2019object}, and object recognition \cite{eitel2015multimodal}, \cite{amirian2018}.
        Deep neural networks are very powerful techniques to extract the generalized feature representation of data. However they require an abundant labeled data for training while annotating large amount of data is laborious, costly and sometimes impossible. Besides, neural networks assume that both source and target domains are sampled from the same distribution, and domain shift can greatly degrade the performance. Hence, deep domain adaptation was proposed to address the lack of sufficient labeled data while boosting the model's performance by deploying deep network properties along with adaptation techniques.\nocite{*} Deep network adaptation techniques are mainly categorized into discrepancy-based, reconstruction-based, and adversarial-based adaptation.
        \subsubsection{Discrepancy-based.}
            For the first time, Long \textit{et al.} \cite{long2015learning}, propose Deep Adaptation Network (DAN) which utilizes the deep neural networks in the domain adaptation setting to learn transferable features across domains. DAN assumes that there is a shift between marginal distributions while the conditional distributions remain unchanged. Therefore, it tends to match the marginal distributions across domains by adding multiple adaptation layers for the task-specific representations. Adaptation layers utilize multiple kernel variant of MMD (MK-MMD) to embed all the task-specific representations into RKHS and align the shift between marginal distributions. DAN only aligns the marginal distributions and does not consider the conditional distribution disparity across domains.
            Deep Transfer Network (DTN) \cite{zhang2015deep}, was proposed to match both marginal and conditional distributions simultaneously. DTN is a MMD-based distribution matching technique and composed of two types of layers. the shared feature extraction layer learns a subspace to align the marginal distributions across domains while discrimination layer matches the conditional distributions using classifier transduction. To reduce the difference in the conditional distributions, the source labels and the target pseudo labels are used to project the data points into different hyperplanes for different classes. 
            To align the conditional distributions, the data points are first projected into different hyperplanes using the source labels and the target pseudo labels and then the discrepancy between conditional distributions is reduced by measuring and minimizing the conditional MMD based objective function.
        \subsubsection{Reconstruction-based}
            Another category of deep adaptation networks utilizes autoencoder to align the discrepancy between domains by minimizing the reconstruction error and learning invariant and transferable representation across domains. The purpose of using autoencoder in domain adaptation is to learn the parameters of the encoder based on the samples in one domain (source) and adapt the decoder to reconstruct the samples in another domain (target).
            Glorot \textit{et al.} \cite{glorot2011domain}, proposed a deep domain adaptation network based on Stacked Auto Encoders (SDA) \cite{vincent2010stacked}, to extract high-level representation to represent both source and target domain data. The proposed model is assumed to have access to various domains with unlabeled data and only one source labeled data. In the first step, the higher-level feature extraction can be obtained by learning from all the available domains in an unsupervised manner.  
            In the first step, the model obtains high-level representations from data in all domains by minimizing the reconstruction error. Next, a linear classifier, such as linear SVM, is trained on the extracted features of the source labeled data by minimizing a squared hinge loss. Finally, this classifier can be used on the target domains. The model shows remarkable results; however, using SDAs make it computationally expensive and unscalable, especially when having high-dimensional features.
            Marginalized SDA (mSDA) \cite{chen2012marginalized}, was proposed to extend the work in SDAs and address its limitations. mSDA marginalizes noise with linear denoisers to induce the model to learn the parameters in a closed-form solution without using stochastic gradient descent (SGD) \cite{bottou2012stochastic}. 
            Ghifary \textit{et al.} \cite{ghifary2016deep}, proposed a Deep Reconstruction-Classification Network (DRCN) that uses an encoder-decoder network for unsupervised domain adaptation in object recognition.
            DRCN consists of a standard convolutional network (encoder) to predict the source labels and a deconvolutional networks \cite{zeiler2010deconvolutional}, (decoder) to reconstruct the target samples. The model jointly utilizes supervised and unsupervised learning strategies to learn the encoder parameters to predict the source labels and the parameters of the decoder to reconstruct the target data. The encoder parameters are shared across both label prediction and reconstruction tasks, while the decoder parameters are private only for the reconstruction task. 
        \subsubsection{Adversarial-based}
            The success of adversarial learning as a powerful domain-invariant feature extractor has motivated many researchers to embed it into deep networks. Adversarial domain adaptation approaches tend to minimize the distribution discrepancy between domains to obtain transferable and domain invariant features. The main idea of adversarial domain adaptation was inspired by the Generative Adversarial Nets (GAN) \cite{goodfellow2014generative}, which tends to minimize the cross-domain discrepancy through an adversarial objective. GANs are deep learning-based generative models composed of a two-player game, a generator model G, and a discriminator model D. The generator aims to produce the samples similar to the domain of interest from the source data, and confuse the discriminator to make a wrong decision. The discriminator then tends to discriminate between the true data in the domain of interest and the counterfeits generated by model G.
        
            Ganin \textit{et al.} \cite{ganin2015unsupervised}, proposed a deep adversarial-based domain adaptation approach to match the domain gap by adding an effective Gradient Reversal Layer (GRL) to the model. During the forward propagation, GRU acts like an identity function that leaves the input unchanged. However, it multiplies the gradient by a negative scalar to reverse it during the backpropagation phase.
            The proposed model can be trained on a massive amount of source labeled data and a large amount of unlabeled target data. The model is generally decomposed into three parts, feature extractor $G_f$, label predictor $G_y$, and domain classifier $G_d$. $G_f$ produces the features for both label predictor and domain classifier. The GRL is inserted between $G_f$ and $G_d$. The three parts of the model can learn their corresponding parameters $\theta_f,$, $\theta_y$, and $\theta_d$ based on the source labeled data and the unlabeled target data.
            During the learning stage, the model learns the parameters $\theta_f$ and $\theta_y$ by minimizing the label predictor loss. This process enforces the feature extractor to produce the discriminative features for a good label prediction on the source domain. At the same time, the model learns the parameters $\theta_f$ and $\theta_d$ by maximizing and minimizing the domain classifier loss, respectively, to obtain domain-invariant features. 
            Pei \textit{et al.} \cite{pei2018multi}, argued that using a single domain discriminator can only match the marginal distributions and ignores the discrepancy between class-conditional distributions causing negative transfer and falsely aligning the classes across domains. Multi-Adversarial Domain Adaptation (MADA)\cite{pei2018multi} was proposed to reduce the shift in the joint distribution between domains and enable fine-grained alignment of different class distributions by constructing multiple class-wise domain discriminators. Such that, each discriminator only matches the source and target data samples belonging to the same class.
        
            The generator G and discriminator D also can be implemented through a minimax optimization of an adversarial objective for a domain confusion. This approach was inspired by theory on domain adaptation \cite{ben2007analysis}, \cite{ben2010theory}, suggesting that the transferable representation across domains does not include discriminative information, and thus, an algorithm is not able to learn the origin of an input sample based on this representation. In other words, the minimax optimization process aims to match the domain disparity by learning deep invariant representation across domains. \cite{tzeng2015simultaneous}, and \cite{ajakan2014domain}, constructed the minimax optimization based adversarial domain adaptation in deep neural networks for the first time.
            Tzeng \textit{et al.} \cite{tzeng2015simultaneous}, introduced an adversarial CNN-based architecture that tends to align both marginal and conditional distributions across domains by minimizing the classification loss, soft label loss, domain classifier loss, and domain confusion loss. The model first computes the average probability distribution, soft label, for the source labeled data in each category. It can be later fine-tuned over the target labeled data to match the class-conditional distributions to the soft labels across domains. Unlike hard labels, the soft labels provide more transferable and useful information about each sample's category and also the relationship between categories, e.g., \emph{bookshelves} are more similar to \emph{filling cabinets} than to \emph{bicycles}.
            The domain confusion loss is the cross-entropy between the uniform distribution over domain labels, and the output predicted domain labels. Minimizing this loss enforces the algorithm to extract invariant representation across domains that maximally confuses the discriminator.
        
            Visual adversarial domain adaptation also utilizes generative adversarial nets (GAN) to reduce the shift between domains. In generative adversarial domain adaptation, generator model G aims to synthesize implausible images, while discriminator model D seeks to identify between synthesized and the real samples. 
            Visual domain adaptation techniques using generative adversarial networks adopt representations in pixel-level, feature-level, or both.
            The pixel-level approaches perform adaptation in the raw pixel space to directly translate the images in one domain into the style images of another domain. 
            Bousmal \textit{et al.} \cite{bousmalis2017unsupervised}, proposed an unsupervised pixel-level domain adaptation method (PixelDA) that utilizes generative adversarial networks to adopt one domain to another. The model changes source domain images to appear as if drawn from the target domain distribution while preserving their original content.
            Simulated+Unsupervised learning (SimGAN) \cite{shrivastava2017learning}, aims to generate synthetic images that are realistic and similar to those in the domain of interest.  A simulator first generates synthetic images from a noise vector. A refiner network then refines them to resemble the target images by optimizing an adversarial loss.
            In contrast to the pixel-level approaches, the feature-level methods modify the representation in the discriminative feature space to alleviate the discrepancy across domains. 
            The Domain Transfer Network (DAN) \cite{taigman2016unsupervised}, is a feature-level generative-based adaptation that enforces the consistency in the embedding space to transforms a source image into a target image.
            Hoffman \textit{et al.} \cite{hoffman2017cycada}, proposed Cycle-Consistent Adversarial Domain Adaptation (CyCADA) that fuses both pixel-level \cite{liu2016coupled}, \cite{bousmalis2017unsupervised}, \cite{shrivastava2017learning}, and feature level \cite{long2015learning}, \cite{tzeng2017adversarial}, adversarial domain adaptation methods with cycle-consistent image-to-image translation techniques \cite{zhu2017unpaired}, to direct the mapping from one domain to another. The model simultaneously adopts representations at both pixel-level and feature-level, by minimizing several losses, including pixel-level, feature-level, semantic, and cycle consistency losses in both domains. 

\bibliographystyle{unsrt}  

\end{document}